\pdfoutput=1
\documentclass[letterpaper, 10 pt, conference]{ieeeconf}  

\IEEEoverridecommandlockouts                              

\usepackage{graphics} 
\usepackage{epsfig} 
\usepackage{subfigure}
\usepackage{mathptmx} 
\usepackage{amsmath} 
\usepackage{booktabs}
\usepackage{threeparttable}
\usepackage{makecell}
\usepackage{setspace}
\usepackage{cite}
\usepackage[T1]{fontenc}
\usepackage[utf8]{inputenc}
\usepackage{authblk}
\usepackage{makecell}
\title{\LARGE \bf
Lane Detection in Low-light Conditions Using an Efficient Data Enhancement : Light Conditions Style Transfer
}

\author{Tong Liu$^{*}$\thanks{$^{*}$Corresponding author: liutong2002@bit.edu.cn}, Zhaowei Chen, Yi Yang, Zehao Wu and Haowei Li
\authorcr{Integrated Navigation \& Intelligent Navigation Laboratory, Beijing Institute of Technology}
}

\begin{document}

\newcommand{\RNum}[1]{\uppercase\expandafter{\romannumeral #1\relax}}
\maketitle
\thispagestyle{empty}
\pagestyle{empty}

\begin{abstract}

Nowadays, deep learning techniques are widely used for lane detection, but application in low-light conditions remains a challenge until this day. Although multi-task learning and contextual-information-based methods have been proposed to solve the problem, they either require additional manual annotations or introduce extra inference overhead  respectively. In this paper, we propose a style-transfer-based data enhancement method, which uses Generative Adversarial Networks (GANs) to generate images in low-light conditions, that increases the environmental adaptability of the lane detector. Our solution consists of three parts: the proposed SIM-CycleGAN, light conditions style transfer and lane detection network. It does not require additional manual annotations nor extra inference overhead. We validated our methods on the lane detection benchmark CULane using ERFNet. Empirically, lane detection model trained using our method demonstrated adaptability in low-light conditions and robustness in complex scenarios. Our code for this paper will be publicly available\footnote{https://github.com/Chenzhaowei13/Light-Condition-Style-Transfer}.

\end{abstract}

\section{INTRODUCTION}
With the development of deep learning, neural networks have been widely used in lane detection\cite{bertozzi1998gold}. Lane detection can be challenging in low-light conditions, such as night, cloudy day and shadows. When the light condition is not ideal, as shown in Fig 1. (a), the lane markings become fuzzy, which makes it difficult to extract features, thus bringing problems to lane detection.
At present, there are two methods for lane detection in low-light conditions: (1) multi-task learning; (2) contextual information. Multi-task learning can provide more monitoring information. For example, VPGNet\cite{Lee_2017_ICCV} adds road marking detection and vanishing point prediction in multi-task learning, which can have a better performance in lane detection. However, this method needs data with additional manual labeling. Contextual information is another method, such as SCNN\cite{pan2018spatial}, replacing layer-to-layer connections with slide-to-slide connections, which exchanges pixel information between rows and columns. Contextual information is helpful for the detector to understand the traffic scene better in low-light conditions, but it introduces extra inference overhead due to the process of the message passing.

\begin{figure}[htbp]
\centering
\subfigure[]{
\begin{minipage}[t]{0.6\linewidth}
\centering
\includegraphics[width=1.6in]{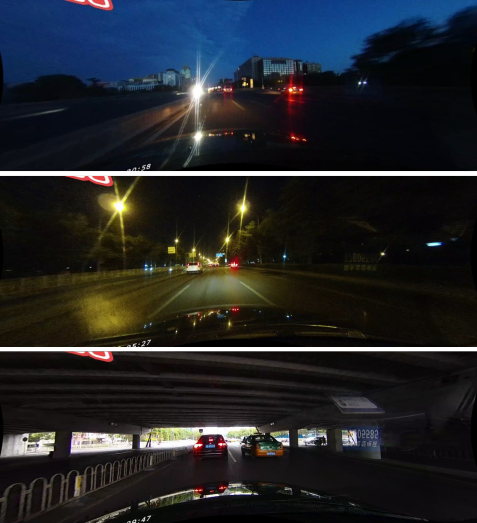}
\end{minipage}%
}%
\subfigure[]{
\begin{minipage}[t]{0.35\linewidth}
\centering
\includegraphics[width=1.2in]{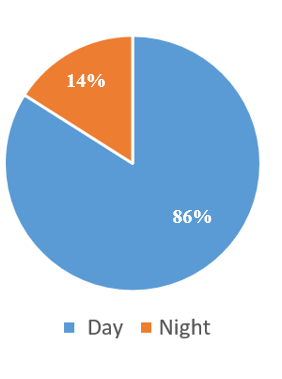}
\end{minipage}
}%
\centering
\caption{(a)Low-light conditions, such as at night and in shadow. (b) Proportion of day and night in CULane training set. The data in low-light conditions make up a very small percentage of the training set.}
\end{figure}
In deep learning, the number of data and the variety of environment directly determine the performance of the detector.  CULane\cite{pan2018spatial} is a challenging lane detection benchmark. Due to the low portion of training data in low-light conditions, the detector performs poorly in these conditions. As shown in Fig 1.(b), in the training set of CULane, the data in night environment only accounts for $14\%$. This is a contributor to the low performance in low-light conditions of CULaned-trained detectors.

In this paper, we propose a style-transfer-based data enhancement method for lane detection, which uses Generative Adversarial Networks(GANs)\cite{goodfellow2014generative} to generate images in low-light conditions and improves the performance of the lane detector.  The proposed light condition style transfer method can be easily applied to various detection tasks and resolve the challenges in low-light conditions. Our method does not require additional data collection and annotation, nor does it increase the inference overhead of the lane detector.

For light conditions style transfer, we propose SIM-CycleGAN, which yields generated images in low-light conditions with higher fidelity comparing to CycleGAN\cite{zhu2017unpaired}. The proposed  SIM-CycleGAN adds scale information match to solve the scale variation problem caused by non-proportional resizing. At the same time, the generated images use the labels from original images, so there is no require for data collection and manual annotation. We put the generated images with the original labels into the training set and feed them to lane detection model for training. In the process of inference, the lane detection model only runs independently without SIM-CycleGAN, which will not introduce extra inference overhead. We use ERFNet\cite{romera2017erfnet} as our lane detection model. In our experiments, the ERFNet trained with our light conditions style transfer method is more robust and performs better than the model without data enhancement not only in low-light conditions, but also in other challenging environments.

Our main contributions are summarized as the following:
\begin{itemize}
\item We propose SIM-CycleGAN with scale information match operation, which solve the scale variation problem and make the generated images more realistic.
\item We propose an efficient data enhancement in low-light conditions using light conditions style transfer method, which does not require additional manual labeling nor extra inference overhead.
\item We validate the effectiveness of proposed method on lane detection model ERFNet.
\end{itemize}

\section{RELATED WORK}

\subsection{Lane Detetion}

Most traditional lane detection methods are based on hand-crafted features. Cheng \textit{et al.}\cite{cheng2006lane} extract lane based on color information and Aly\cite{aly2008real} apply  Line Segment Detection (LSD), which is followed by post-processing steps. Apart from color features and gradient feature, ridge features are also applied in lane detection, such as Hough transform\cite{jung2013efficient}, Kalman filtering\cite{kim2008robust} and Particle filtering\cite{borkar2011novel}.
Song\cite{Song2018Lane} design a self-adaptive traffic lanes model in Hough Space with a maximum likelihood for lane detection.

Since Convolutional Neural Network (CNN) can extract features at different levels, it has been widely used for lane detection tasks in different scenes\cite{neven2018towards, ghafoorian2018gan}. Van Gansbeke \textit{et al.}\cite{de2019end} proposed a lane detector in an end-to-end manner, which consists of two parts: a deep network that predicts a segmentation-like weights map for each lane, and a differentiable least-squares fitting module for returning fitting parameters. Y.HOU \textit{et al.}\cite{hou2019learning} present Self Attention Distillation(SAD) for lane detection, while further representational learning is achieved by performing top-down and hierarchical attention distillation networks within the network itself.

\subsection{GANs for Image Transfer}

Based on Conditional of Generative Adversarial Networks (cGANs)\cite{mirza2014conditional}, there are some methods for supervised image-to-image translation between unpaired samples\cite{zhu2017unpaired, Isola_2017_CVPR}. CycleGAN\cite{zhu2017unpaired} is essentially two mirrored and symmetrical GANs, which share two generators and each carries a discriminator, enabling unsupervised transfer between the two fields. Compared with CycleGAN, the proposed SIM-CycleGAN avoids non-proportional resizing and generates more realistic images.

\begin{figure*}
\centering
\begin{minipage}[!htbp]{1\linewidth}
\includegraphics[width=7in]{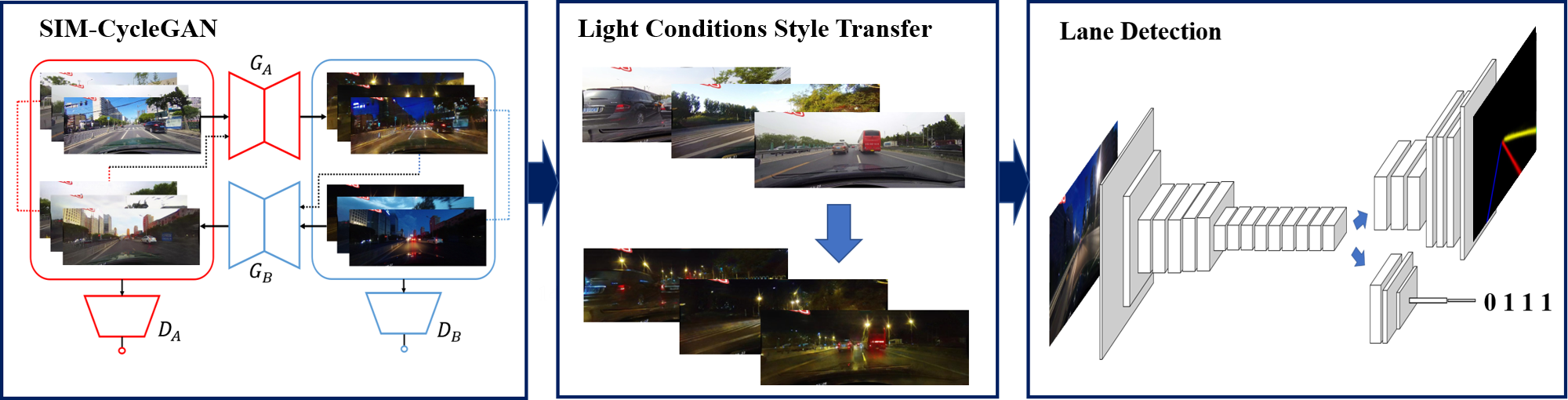}
 \end{minipage}
 \caption{The main framework of our method. The proposed SIM-CycleGAN is shown on the left. The generator $G_{A}$ transfer images from suitable light conditions to low-light conditions, while the generator $G_{B}$ transfer in the opposite way. The discriminator $D_{A}$ and $D_{B}$ feed the single scalar value(real or fake) back to generators. The middle section shows light conditions style transfer from suitable light conditions to low-light conditions by the trained SIM-CycleGAN. Lane detection model is shown on the right, whose baseline is ERFNet. We add lane exist branch for better performance. }
 \end{figure*}

\subsection{GANs for Data Enhancement}

There are works using GANs to generate images for data enhancement to improve object detection. Xiaolong W \textit{et al.}\cite{wang2017fast} consider that object occlusion or deformation account for a small percentage, so Adversarial Spatial Dropout Network(ASDN) and Adversarial Spatial Transformer Network(ASTN) are designed for learning how to occlude and rotate object respectively. Lanlan Liu \textit{et al.}\cite{liu2019generative} develop a model that jointly optimizes GANs and detector. It is trained by feeding the generator with detection loss, which can generate objects that are difficult to detect by the detector and improve the robustness of the detector. The methods above only use GANs to solve problems of object occlusion and small object detection, but it has no effect on the problem of low-light conditions.

Closely related to our work, Vinicius F. \textit{et al.}\cite{arruda2019cross} complete the style transfer from day to night through CycleGAN to enhance the data in night scenes. Nevertheless, when using CycleGAN for style transfer, the author does not consider the inconsistent width and height of the images taken by the vehicle camera, directly cut out the two sides of the images and then resize the images to $256\times256$ , which ignores the challenging detection areas. In contrast, the proposed method adds scale information match on generator to avoid forcing resize, which can better ensure the authenticity of the generated images and has certain benefits for the training of the detector.

\section{PROPOSED METHOD}

The proposed method, illustrated in Fig 2, consists of three steps: (i) SIM-CycleGAN, (ii) light conditions style transfer, and (iii) lane detection. SIM-CycleGAN is composed of two generators and two discriminators, when generators focus on generating images and discriminators give feedback to the generator about whether the generated images are authentic. We use SIM-CycleGAN to realize light conditions style transfer from suitable light conditions to low-light conditions, in order to implement data enhancement without any manual collection and labeling. Finally, we use ERFNet as our lane detection model. ERFNet is trained with the data after enhancement by light conditions style transfer, and it infers by itself without any other process.

\subsection{SIM-CycleGAN}

\begin{figure}[thpb]
\centering
\includegraphics[scale=0.40]{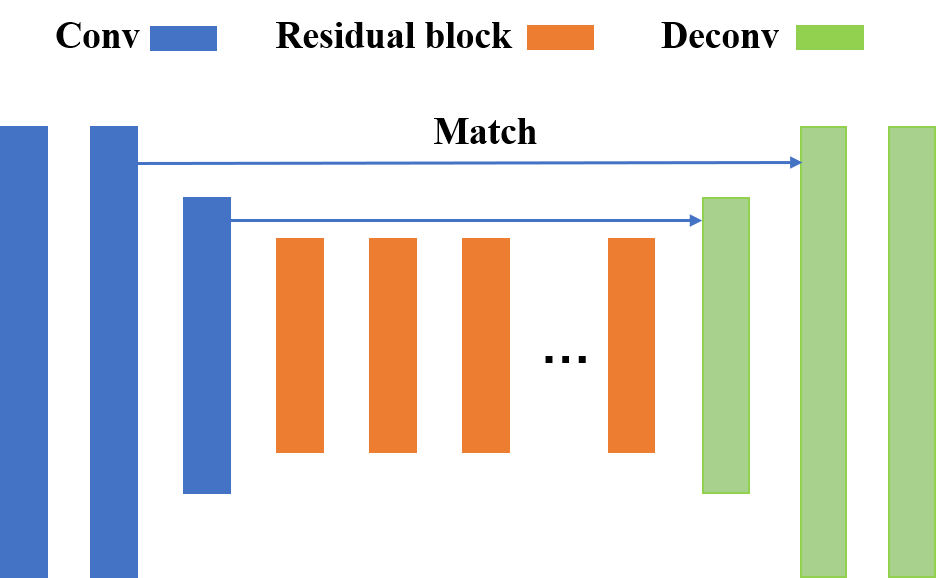}
\label{figurelabel}
\caption{Generator architecture, composed of convolution layers, residual blocks and deconvolution layers. Convolution layers record the changing information of scale in the encoding process and maps it to the corresponding operation in the decoding process. }
\end{figure}

To accommodate images with different sizes from different datasets, we need preprocess including resizing and cropping before we send images to network, which could easily lead to the distortion and information loss in the generated images. SIM-CycleGAN can achieve automatic scale adaptation by adding scale information match operation on generator, making the generated images more realistic. We implement our architecture based on CycleGAN, which is composed of two generators and two discriminators.

\textbf{Generator Network} Inspired by the architecture of the CycleGAN, the generator of SIM-CycleGAN contains convolution layers, residual blocks and deconvolution layers, as shown in Fig 3. As we known, the convolutional auto-encoder cannot be applied to different resolution images. However, if we remove the preprocess steps, inputs should add padding to required resolution (multiple of 4) for the convolutional auto-encoder, which make the resolution of generated images being different from the original image. Scale information match of SIM-CycleGAN automatically records the change information of scale in the encoding process and match it to the corresponding operation in the decoding process. This operation can solve the scale variation problem caused by non-proportional resizing and realize adapting different resolution images without any other process.

\textbf{Discriminator Network } For the discriminator network we use PatchGAN\cite{Isola_2017_CVPR} instead of full-image discriminator, which can classify whether $70\times70$ overlapping patches are real or fake with fewer parameters. Discriminator outputs the result of judging the generated image and feed it back to the generator.

\subsection{Light Conditions Style Transfer}

In this paper, light conditions style transfer is completed by SIM-CycleGAN. We define the set of images with suitable light conditions as domain X, and the set of images with low-light conditions as domain Y. SIM-CycleGAN can transform images from domain X to domain Y and also domain Y to domain X.

In training process, the generator generates more realistic images based on the feedback from the discriminator, which aims to achieve the purpose of spoofing discriminator. The discriminator judges the authenticity (real/fake) of the generated images more accurately, and finally achieves dynamic balance. Therefore, we introduce adversarial loss to describe this process, which is the key of GANs to generate realistic images. We define the adversarial loss from domain X to domain Y as:
\begin{align}
L_{GAN}(G_{A},D_{B},X,Y) & =  E_{y\sim P_{data}(y)}[logD_{B}(y)] \nonumber \\
& + E_{x\sim P_{data}(x)}[log(1-D_{B}(G_{A}(x)))]
\end{align}
where $G_{A}$ tries to generate images $G_{A}(x)$ that look like images from domain Y by minimizing this objective, while $D_{B}$ aims to distinguish between generated samples $G_{A}(x)$ and real samples y by maximizing this objective, i.e. $min_{G_{A}}max_{D_{B}}L_{GAN}(G_{A},D_{B},X,Y)$. The adversarial loss from domain Y to domain X is similar to Eqa. 1: $min_{G_{B}}max_{D_{A}}L_{GAN}(G_{B},D_{A},Y,X)$.

For better performance, we adapt cycle-consistent like $x \to G_{A}(x) \to G_{B}(G_{A}(x)) \approx x$ and $y \to G_{B}(y) \to G_{A}(G_{B}(y)) \approx y$. This function monitors the difference between the original images and the generated images after style transferred to and from the opposite domain, thus leading the generated images to be more realistic. We incentivize this behavior using a cycle consistency loss:
\begin{align}
L_{cyc}(G_{A},G_{B}) & =  E_{x\sim P_{data}(x)}[\lVert G_{B}(G_{A}(x))-x \rVert _{1}] \nonumber \\
& + E_{y\sim P_{data}(y)}[\lVert G_{A}(G_{B}(y))-y \rVert _{1}]
\end{align}

The total loss is the sum of the three items.
\begin{align}
L(G_{A},G_{B},D_{A},D_{B}) &=  L_{GAN}(G_{A},D_{B},X,Y) \nonumber \\
&  + L_{GAN}(G_{B},D_{A},Y,X) \nonumber \\
& + \lambda L_{cyc}(G_{A},G_{B})
\end{align}
where $\lambda $ controls the relative importance of the two objectives.

Some examples of light conditions style transfer result are shown in Fig 4. The generated images use the annotations from corresponding original images, which doesn't require additional annotations.
It is convenient to drop SIM-CycleGAN for data enhancement into existing workflow.

 \begin{figure}[thpb]
\centering
\includegraphics[scale=0.8]{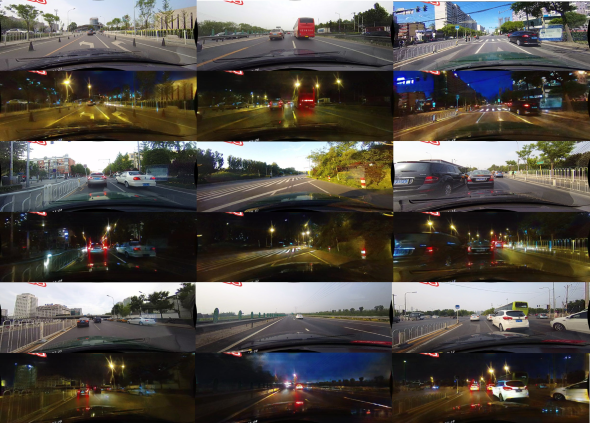}
\label{figurelabel}
\caption{Examples of real images in normal light conditions and their corresponding style-transferred low-light condition counterparts. Although some details are not well handled, most of generated images can be converted to low-light conditions with high fidelity.}
\end{figure}

\subsection{Lane Detection}

For lane detection, we use ERFNet as our baseline, which has a novel layer that uses residual connections and factorized convolutions in order to remain efficient while retaining remarkable accuracy. Considering slow convergence of ERFNet, we add lane existence branch as shown in Fig 5. In our architecture, decoder takes in charge of the instance segmentation task, which outputs probability maps of a set of lane markings per pixel. The second branch outputs the confidence of corresponding lane markings. The loss function is shown as follow:
\begin{align}
Loss =  \lambda_{1} L_{seg} + \lambda_{2} L_{exist}
\end{align}
where $L_{seg}$ is instance segmentation negative log likelihood loss, $L_{exist} $ is lane existence binary cross entropy loss. We balance the tasks by weight terms $\lambda_{1}$, $ \lambda_{2}$ and finally set $ \lambda_{1}=0.9$, $ \lambda_{2}=0.1$.

During training, the original images and generated images are fed to ERFNet. In inference time, we just need ERFNet without any other operation and then draw lanes from the output probability maps. For each lane marking with a confidence value larger than 0.5, we search the corresponding probability map every 20 rows for the position with the highest response. In the end, these positions are then connected by lines, which are the final predictions.

\begin{figure}[thpb]
\centering
\includegraphics[scale=0.40]{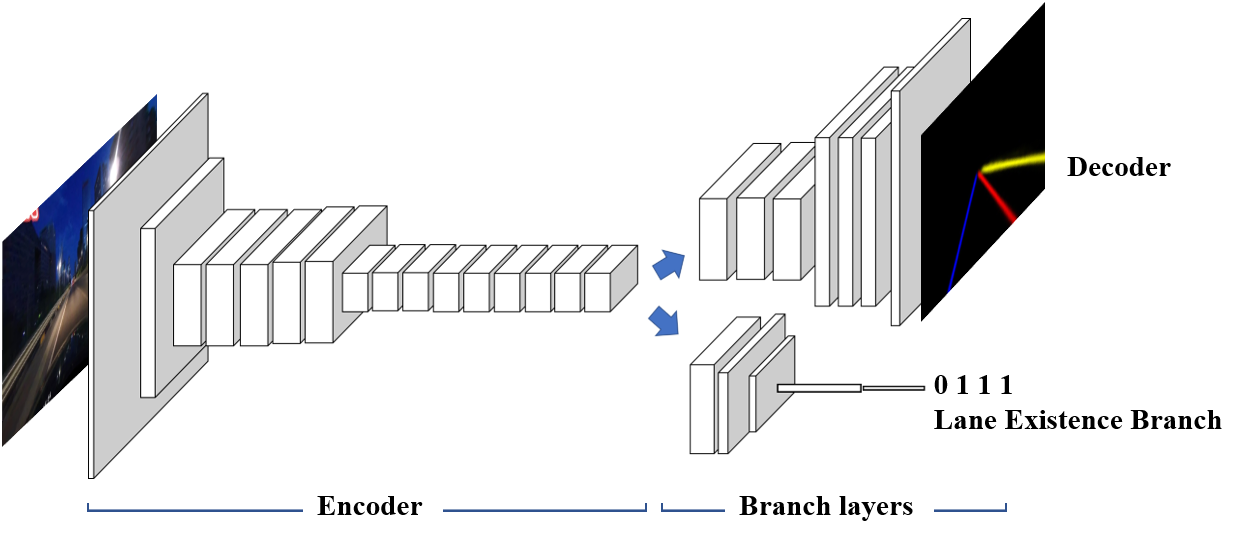}
\caption{Our lane detection model architecture. The decoder outputs probability maps of different lane markings, and the second branch predicts the existence of lane.  }
\label{figurelabel}
\end{figure}

\begin{table*}
  \centering
  \small
  \caption{Performance ($F_{1}$-measure) of different methods on CULane testing set. For crossroad, only FP is shown. }\label{label:1}
  \begin{tabular}{p{2cm}<{\centering}|cp{1.8cm}<{\centering}p{2.8cm}<{\centering}|p{1.5cm}<{\centering}p{2.3cm}<{\centering}p{2.5cm}<{\centering}}
        \hline
        Category & ERFNet  & CycleGAN + ERFNet & SIM-CycleGAN + ERFNet(ours) & SCNN\cite{pan2018spatial} & ENet-SAD\cite{hou2019learning} &ResNet-101-SAD\cite{hou2019learning}   \\
        \hline
        Normal   & 91.5   & 91.7 & \bf91.8 & 90.6 & 90.1 & 90.7                    \\
        Crowded & 71.6   & 71.5 & \bf71.8  & 69.7 & 68.8 & 70.0                  \\
        Night & 67.1   & 68.9 & \bf69.4  & 66.1 & 66.0 & 66.3                  \\
        No Line & 45.1  & 45.2 & \bf46.1   & 43.4 & 41.6 & 43.5                  \\
        Shadow  & 71.3   & 73.1 & \bf76.2  & 66.9 & 65.9 & 67.0                 \\
        Arrow   & 87.2   & 87.2 & \bf87.8  & 84.1 & 84.0 & 84.4                 \\
        Dazzle Light & 66.0 & \bf67.5& 66.4& 58.5 & 60.2 & 59.9                    \\
        Curve & 66.3  & \bf69.0 & 67.1     & 64.4 & 65.7 & 65.7              \\
        Crossroad & 2199  & 2402 & 2346 & \bf1990 & 1998 & 2052                  \\
        \hline
        Total & 73.1  & 73.6 & \bf73.9     & 71.6 & 70.8 & 71.8               \\
        \hline
  \end{tabular}
\end{table*}

 \begin{figure*}
\centering
\begin{minipage}[!htbp]{0.99\linewidth}
\includegraphics[width=6.9in]{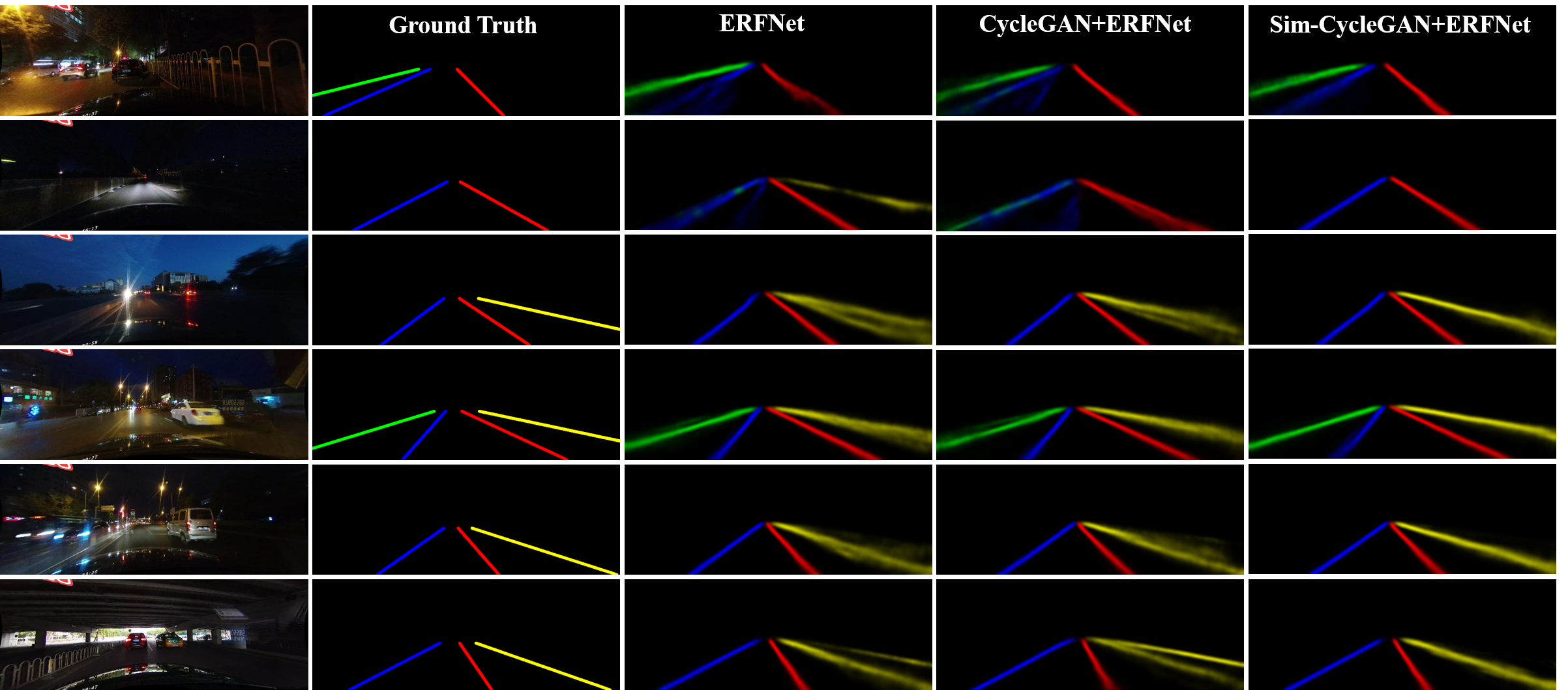}
 \end{minipage}
 \caption{The probability maps from our method and other methods. The brightness of the
pixel indicates the probability of this pixel belonging to lanes. It can be clearly seen from this figure, in low-light conditions, the probability maps generated by our method is more pronounced and more accurate.}
 \end{figure*}

\section{EXPERIMENTAL RESULTS}
\subsection{Dataset}
CULane\cite{pan2018spatial} dataset is widely used in lane detection, which contains various scenario that’s challenging for driving, like crowded, shadow, night, dazzle light and so on.

To train the light-condition style transfer model, 3200 images in suitable light conditions and 3,200 images in low-light conditions have been selected. These images are divided by a 3:1 ratio, which are respectively used as the training set and the test set of SIM-CycleGAN. After finishing the training of SIM-CycleGAN, we select 13,000 images in suitable light conditions to be transferred into in low-light conditions through SIM-CycleGAN.

\subsection{Implementation Details}

\textbf{Light conditions style transfer }
Due to the limited memory of GPU, we resize images to $820\times295$. SIM-CycleGAN is trained with 100 epochs, with one image per batch. In order to compare the effects of SIM-CycleGAN with CycleGAN, we also do similar work on CycleGAN, and resize images to $256\times256$ before training.

\textbf{Lane detection}
A public source code\footnote{https://github.com/cardwing/Codes-for-Lane-Detection/tree/master/ ERFNet-CULane-PyTorch} is used for carrying out the experiments. Before training, we resize the images in CULane to $820\times295$. ERFNet is pre-trained by Cityscape dataset. Our lane detection model is trained with 12 epochs, with 12 images per batch. We use SGD to train our models and the initial learning rate is set to 0.01.

In order to verify the effectiveness of our method, we design a cogent comparative experiment, which includes three groups: (1) the original CULane training set, (2) CULane training set adding 13,000 images in low-light conditions generated by CycleGAN, and (3) CULane training set adding 13,000 images in low-light conditions generated by SIM-CycleGAN. These three groups are named as ERFNet, ERFNet+CycleGAN and ERFNet+SIM-CycleGAN respectively. The test set of the experiment is CULane test set.

\subsection{Evaluation metrics}

Following [3], in order to judge whether a lane is correctly detected, we treat each lane marking as a line with 30 pixel width and compute the intersection-over-union (IoU) between labels and predictions. Predictions whose IoUs are larger than threshold are considered as true positives (TP). Here threshold is set to 0.5. Then, we use F1 measure as the evaluation metric, which is defined as: $F_{1}= \frac{2\times Precision \times Recall}{Precision+Recall}$, where $Precision = \frac{TP}{TP+FP}$, and $Recall = \frac {TP}{TP+FN}$.

\subsection{Results}

The results of our comparative experiment are shown in Table \textrm{I}. Compared with ERFNet without data enhancement, we can find that our method (SIM-CycleGAN+ERFNet) perform better in low-light conditions, such as night and shadow, whose $F_{1}$ measure improves $2.3\%$ and $4.9\%$ respectively. This indicates that the light conditions style transfer method is helpful to the lane detection performance of ERFNet in low-light conditions. At the same time, light conditions style transfer also helps our lane detection model perform better in other scenarios, which increase the total $F_{1}$ measure from $73.1\%$ to $73.9\%$. Specifically, our method improves the $F_{1}$ measure from $45.1\%$ to $46.1\%$ in no line scenarios and from $66.0\%$ to $66.4\%$ in dazzle light scenarios. It shows that our method not merely can help ERFNet achieve lane detection better in low-light conditions, but also prompt ERFNet to understand different lane markings in other challenging scenarios.

We also add the CycleGAN+ERFNet method into the comparison experiment. The result shows that the proposed method is also superior to CycleGAN+ERFNet in most traffic scenes. Because the low-light conditions images generated by SIM-CycleGAN is more realistic than the images generated by CycleGAN, it eliminates the negative impact on training from low quality generated images on the network.

A sample of the probability maps outputted by the three methods are shown in the Fig 6. The probability maps generated by our method are more pronounced and more accurate, which comes to the conclusion that our method can extracts the characteristics of lane markings better in low-light conditions.

\subsection{Ablation Study}
\textbf{Generated Images v.s. Real Images }
For comparing influence of the generated images with real images, 13,000 images in low-light conditions from the original CULane training set and 13,000 images in low-light conditions generated by SIM-CycleGAN are used as the training set. The CULane validation set is used for validation. The results are shown as Fig. 7, which indicate that the model using generated images could converge and overfit faster. Because the data collected in the real environments will have a lot of interference and noise that cannot be eliminated, while the images generated by SIM-CycleGAN can be easily understood by lane detection model. In addition, since deep learning is a probability distribution problem, the images generated by SIM-CycleGAN can be closer to the distribution of the actual traffic scenarios, so that the data-driven detection model can understand different lane markings better.


\textbf{Image Amounts to Generate} We fine tune the number of generated images for our lane detection model. Assume that the ratio of the generated images to the real image in low-light conditions is N:1,  we take $N = 0.25, 0.5, 1, 2, 4 $ for comparative experiment.
The results are shown as Table \textrm{II}. Although having more generated images in low-light conditions, the $F_{1}$-measure of $N=2,4$ shows no advantage over experiments with smaller N, and performs worse in other scenes.
Because too much images in low-light conditions don't coincide the distribution of the actual traffic scenarios, which leads to the trained model that prefers low-light conditions and performs poorly in other scenarios.
In this experiment, $N=1$ is the best ratio to generate images by light conditions style transfer.
In conclusion, the appropriate amounts of generate images is beneficial to optimized performance on lane detection.

\begin{figure}
\centering
\includegraphics[scale=0.55]{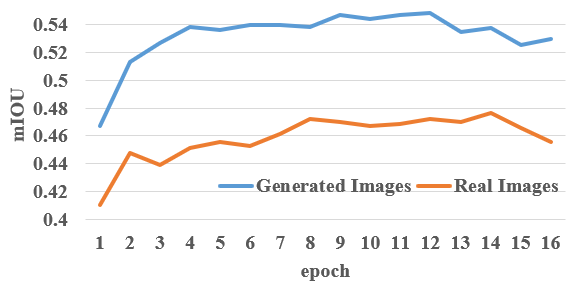}
\caption{The training process of ERFNet on real images and generated images. The ordinate is mIOU of validation set, and the abscissa is epoch. }
\label{figurelabel}
\end{figure}

\begin{table}
  \centering
  \small
  \caption{ $F_{1}$-measure of different N on CULane testing set. For crossroad, only FP is shown. }\label{label:1}
  \begin{tabular}{c|ccccc}
        \hline
        Category & N=0.25  & N=0.5 & N=1 & N=2 & N=4  \\
        \hline
        Normal   & 91.8    & \bf91.9 & 91.8 & 91.7 & 91.5                    \\
        Crowded & 71.5    & 72.0  & 71.8 & 71.9& \bf72.3                     \\
        Night & 68.7    & 69.6 & 69.4 & \bf69.7 & \bf69.7                     \\
        No Line & 45.7    & 46.1 &46.1 & \bf46.9 & 46.6                   \\
        Shadow  & 70.1    & 71.5 &\bf76.2 & 71.6  & 75.3                   \\
        Arrow   & 87.1   &  87.6  &\bf87.8 & 87.6 & 87.2                 \\
        Dazzle Light & 65.6 & 64.4 &\bf66.4 & 64.8  & 65.3                   \\
        Curve & 68.3    & \bf68.8    &67.1 & 68.4 & 68.5                  \\
        Crossroad & 2385   & 2739 & \bf2346  & 2622 & 2738                   \\
        \hline
        Total & 73.5    & 73.7 &  \bf73.9 & 73.8 & 73.8                 \\
        \hline
  \end{tabular}
\end{table}

\section{CONCLUSIONS}

Lane detection can be a challenge in low-light conditions. In this paper, we propose a style-transfer-based data enhancement method for lane detection in low-light conditions. Our method uses the proposed SIM-CycleGAN to generate images in low-light conditions for improving the environmental adaptability for lane detector. Our workflow does not require additional annotations nor extra inference overhead.
We have validated our method on CULane, in which our technique achieves better lane detection results not only in low-light conditions but also in other challenging scenarios.
Since we do not jointly train SIM-CycleGAN and ERFNet, it remains unexplored to build an end-to-end system where the style-transfer would generated images under various challenging scenarios that would further improve the performance of lane detector in a wider spectrum.
We would like to explore this direction in future work.

\section{ACKNOWLEDGMENT}
This work was partly supported by National Natural Science Foundation of China (Grant No. NSFC 61473042).

\addtolength{\textheight}{-12cm}   



\bibliographystyle{IEEEtran}
\bibliography{a}

\end{document}